# Causal Inference in the Presence of Latent Variables and Selection Bias

**Peter Spirtes, Christopher Meek, and Thomas Richardson**
Department of Philosophy
Carnegie Mellon University
Pittsburgh, PA 15213

## Abstract

We show that there is a general, informative and reliable procedure for discovering causal relations when, for all the investigator knows, both latent variables and selection bias may be at work. Given information about conditional independence and dependence relations between measured variables, even when latent variables and selection bias may be present, there are sufficient conditions for reliably concluding that there is a causal path from one variable to another, and sufficient conditions for reliably concluding when no such causal path exists.

## 1  INTRODUCTION

There are well known problems with drawing causal inferences from samples that have not been randomly selected. Spirtes et al. (1993) showed that otherwise correct discovery algorithms fail even in the large sample limit when samples are selected on the basis of features with certain causal connections to the variables under study. For instance, if X and Y are independent in a population, but a sample is selected from the population using some value of a variable Z that happens to be influenced by X and Y, then X and Y will have a statistical dependency in the sample (produced by conditioning on Z), that they do not have in the population. Cooper (1995) has given a number of more interesting examples of this kind involving latent variables. Methods of representing selection bias, and special cases where selection bias is detectable from data were discussed in Wermuth, Cox, and Pearl (1994). An important question is whether there are any general, informative and reliable procedures for discovering causal relations when, for all the investigator knows, both latent variables and selection bias may be at work. When selection bias does not apply, there is an algorithm, FCI, that under assumptions described below will (in the large sample limit) almost certainly give such information including information about the existence or non-existence of causal path-ways from one measured variable to another (Spirtes et al. 1993). We have shown that under a reinterpretation of the output, the FCI algorithm also applies when selection bias may be present, but the output is then generally less informative than in cases known to be free of selection bias. We have also shown, however, that given information about conditional independence and dependence relations between measured variables, even when latent variables and selection bias may be present, there are sufficient conditions for reliably concluding that there is a causal path from one variable to another, and sufficient conditions for reliably concluding when no such causal path exists.

Throughout this paper, sets of variables are in boldface, and defined terms in italics. Graph theoretic terms are defined in the appendix. The FCI algorithm and proofs of its correctness are described in detail in Spirtes et al. (1993). Due to lack of space we will not describe the algorithm here. In the main body of this paper we focus on explaining the adaptations and reinterpretations that selection bias requires for the interpretation of the output. The theorems given in the Appendix to this paper are proved by straightforward if tedious adaptations to the selection bias case of the proofs, given in Spirtes et al. (1993), of the correctness of the FCI algorithm without selection bias.

## 2  DIRECTED ACYCLIC GRAPHS

Factor analysis models, path models with independent errors, recursive linear structural equation models with independent errors, and various kinds of latent variable models are all instances of directed acyclic graph models. A *directed acyclic graph* (DAG) G with a set of vertices **V** can be given both a causal interpretation and a statistical interpretation. (See Pearl 1988, where under the statistical interpretation DAG models are called Bayesian networks; Spirtes et al. 1993; Wright 1934.) Take "A is a direct cause of B in a member of the population with respect to a set of variables **V**" as primitive. If **V** is a set of variables, there is an edge from A to B in a causal DAG G for a population with variables **V** if and only if A is a *direct cause* of B relative to **V** for some member of that pop-ulation. A DAG G with a set of vertices **V** can also represent a set of probability measures over **V** subject to restrictions relating allowable measures on **V** to the graphical structure of G. Following the terminology of Lauritzen et al. 1990 we say that a probability measure over a set of variables **V** satisfies the *local directed Markov property* for a DAG G with vertices **V** if and only if for every W in **V**, W is independent of the set of all its non-parental non-descendants conditional on the set of its parents.

# 3    REPRESENTATION OF SELECTION BIAS

We distinguish two different reasons why a sample distribution may differ from the population distribution from which it is drawn. The first is simply the familiar phenomenon of sample variation, or as we shall say, *sample bias*: for a given population distribution, the parameter estimates made from a finite random sample of variables do not in general exactly equal the population parameters. The second reason is that causal relationships between variables in **V**, on the one hand, and the mechanism by which individuals in the sample are selected from a population, on the other hand, may lead to differences between the expected parameter values in a sample and the population parameter values. In this case we will say that the differences are due to *selection bias*. Sampling bias tends to be remedied by drawing larger samples; selection bias does not. We will not consider the problems of sample bias in this paper; we will always assume that we are dealing with an idealized selected subpopulation of infinite size, but one which may be selection biased.

For the purposes of representing selection bias, following Cooper (1995) we assume that for each measured random variable A, there is a binary random variable $S_A$ that is equal to one if the value of A has been recorded for that member of the population, and is equal to zero otherwise. If **V** is a set of variables, we will always suppose that **V** can be partitioned into three sets: the set **O** (standing for observed) of measured variables, the set **S** (standing for selection) of selection variables for **O**, and the remaining variables **L** (standing for latent). In the marginal distribution over a subset **X** of **O** in a selected subpopulation, the set of selection variables **S** has been conditioned on, since its value is always equal to 1 in the selected subpopulation. Hence for disjoint subsets **X**, **Y**, and **Z** of **O**, we will assume that we cannot determine whether $X \perp\!\!\!\perp Z | Y$, but that we can determine whether $X \perp\!\!\!\perp Z | Y \cup (S = 1)$. ($X \perp\!\!\!\perp Z | Y$ means X is independent of **Z** given **Y**. If **Y** is empty, we simply write $X \perp\!\!\!\perp Z$. If the only member of **X** is X, then we write $X \perp\!\!\!\perp Z | Y$ instead of $\{X\} \perp\!\!\!\perp Z | Y$.) There may be cases in which all of the variables in **S** always take on the same value; in such cases we will represent the selection with a single variable S.

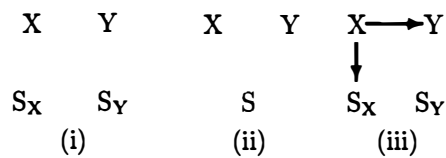

Figure 1: Three causal DAGs

The three causal DAGs for a given population shown in Figure 1 illustrate a number of different ways in which selection variables can be related to non-selection variables. The causal DAG in (i) would occur, for example, if the members of the population whose X values were recorded and the members of the population whose Y values were recorded were randomly selected by flips of a pair of independent coins. The DAG in (ii) would occur if the flip of a single coin was used to choose units which would have both their X and Y values recorded (i.e. $S_X = S_Y$ and there are no missing values in the sample). The DAG in (iii) would occur if, for example, X is years of education, and people with higher X values respond to a questionnaire about their education — and thus appear in the sample — more often than people with lower X values. We do not preclude the possibility that a variable Y can be a cause of $S_X$ for some variable $X \neq Y$, nor do we preclude the possibility that $S_X$ can be a cause of two different variables.

To draw correct causal conclusions about the unselected subpopulation which we have not seen, or about the whole population part of which we have not seen, some assumptions must be made. First, consider the case where one is interested in causal inferences about the whole population from the selected subpopulation. The notion of a causal graph, as we have defined it is relative to a set of variables and a population. Hence the causal graph of the whole population and the causal graph of the selected subpopulation may be different. For example, if a drug has no effect on survival in men, but it does have an effect on women, then there is an edge from drug to survival in the causal graph of the population, but no edge from drug to survival in the causal graph of a subpopulation of men. Because of this, in order to draw causal conclusions about either the population or the unselected subpopulation from the causal graph of the selected subpopulation, we will make the following assumption:

**Population Inference Assumption**: If **V** is a set of variables, then the causal graph over **V** of the population is identical with the causal graphs over **V** of the selected subpopulation and the unselected subpopulation.

We make several additional assumptions relating probability distributions to causal relationships which we introduce with the following example. The first principle simply states that in a causal DAG each variable is independent of its non-descendants (i.e. the variables it does not affect even indirectly) given its parents (i.e.

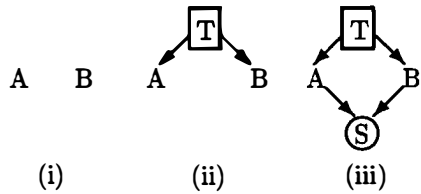

Figure 2: Representing selection bias

its direct causes). However, it is important to realize that this principle does not apply to arbitrary sets of variables or to arbitrary subpopulations, as the following example shows. Let Pop' be a subpopulation of a population Pop, such that every member of Pop' is assigned the value $\mathbf{S} = \mathbf{1}$. Suppose that (i) and (ii) in Figure 2 are correct causal DAGs for Pop' (although (i) is more detailed), and that (iii) of Figure 2 is a correct causal DAG for Pop. (In figures we place latent variables in boxes and selection variables in circles.) Also suppose that A and B are dependent in both Pop' and Pop, but that A and B are dependent given T in Pop' but not in Pop.

(i) of Figure 2 is a correct causal DAG because A is not a direct cause of B and B is not a direct cause of A. Note however, that it is incomplete, because it does not show that there is a latent common cause of A and B. Moreover, it is so incomplete that the local directed Markov Property does not hold for (i) because A and B are dependent in Pop'. DAG (ii) of Figure 2 is a more detailed picture of the causal relations between A and B. However, it is not the case that the causal DAG for the set of variables {A,B,T} satisfies the local directed Markov Property for (ii) in Pop', because A and B are dependent given T in Pop'. (iii) of Figure 2 is a more detailed causal DAG than either (i) or (ii). Moreover, {A,B,T,S} does satisfy the local directed Markov Property for (iii) of Figure 2 because in the population Pop (i.e. where we do not condition on S), A and B are independent given T. So by including latent variables and expanding the population from Pop' to Pop, we eventually reach a point where we find a population and a distribution over the variables in that population satisfying the local directed Markov Property for the causal DAG for that set of variables.

**Causal Markov Assumption:** For each population Pop' and set of variables $\mathbf{V}'$, there is a population Pop from which Pop' is selected by $\mathbf{S}$, a set $\mathbf{V} \supseteq \mathbf{V}' \cup \mathbf{S}$, a causal graph G of Pop over $\mathbf{V}$, and a distribution $P(\mathbf{V})$ in Pop that satisfies the local directed Markov property for G. (We call such a set $\mathbf{V}$ *causally sufficient* for Pop', Pop, and $\mathbf{S}$.)

The Causal Markov Assumption is entailed by many causal models commonly found in the statistical literature, including all recursive structural equation model with independent errors. (For an introduction to linear recursive structural equation models see Bollen 1989.) The Causal Markov Assumption does not generally hold when there are causal interactions between different members of the population (as opposed to causal interactions among properties of individual members of the population), nor does it generally hold in populations with feedback.

The Causal Faithfulness Assumption basically states that any conditional independence relation true in the subpopulation is true for structural reasons (i.e. because of the DAG structure), rather than because of the particular parameterization of the DAG.

**Causal Faithfulness Assumption:** If Pop' is a subpopulation of Pop selected by $\mathbf{S}$, $\mathbf{V}$ is a set of causally sufficient variables, G is a causal graph for population Pop with variables $\mathbf{V}$, and $P(\mathbf{V})$ is the distribution of $\mathbf{V}$ in Pop, then for any disjoint $\mathbf{X}$, $\mathbf{Y}$, and $\mathbf{Z}$ in $\mathbf{V}$, $\mathbf{X} \perp\!\!\!\perp \mathbf{Z} | \mathbf{Y} \cup (\mathbf{S} = 1)$ only if the Causal Markov Assumption and $\mathbf{S} = 1$ entails $\mathbf{X} \perp\!\!\!\perp \mathbf{Z} | \mathbf{Y} \cup (\mathbf{S} = 1)$. (In this case we say that $P(\mathbf{V})$ is faithful to G in Pop' selected by $\mathbf{S}$.)

The Causal Faithfulness Condition may fail for a number of reasons. It could be that some conditional independence relation holds for some particular parameterizations of the DAG and not others, rather than because of the graphical structure. It may also be violated when there are deterministic relationships between variables. However, Spirtes et al. (1993) and Meek (1995) have shown that under natural parameterizations of the linear normal structural equation models and discrete Bayesian networks, the set of parameterizations that lead to violations of faithfulness have Lebesgue measure 0.

The justification of axioms similar to these (without the restriction to conditioning on $\mathbf{S}$) is discussed in Spirtes et al. (1993). As shown in subsequent sections, these assumptions will allow us to draw reliable inferences about the causal graph of the population from the conditional independence relations in the selected subpopulation. If a conditional independence relation is true in every distribution that satisfies the local directed Markov property for DAG G, we say that G *entails the conditional independence relation*; similarly, if a conditional dependence relation is true in every distribution that is faithful to DAG G, we say that G *entails the conditional dependence relation*.

## 4    EXAMPLES

We now consider several different sets of conditional independence and dependence relations, and what they can tell us about the causal DAGs that generated them, under a variety of different assumptions.[1]

Given a causal graph G over a set of variables $\mathbf{V}$, we will say there is no selection bias if and only if for any three disjoint sets of variables $\mathbf{X}$, $\mathbf{Y}$, and $\mathbf{Z}$ included in $\mathbf{V}\backslash\mathbf{S}$, G entails $\mathbf{X}\perp\!\!\!\perp\mathbf{Z}|\mathbf{Y}\cup\mathbf{S}$ if and only if G entails $\mathbf{X}\perp\!\!\!\perp\mathbf{Z}|\mathbf{Y}$. (Note that this does not in general entail that the *distributions* in the selected subpopulation and the population are the same; it just entails that the same conditional independence relations hold in both.) This happens, for example, when the variables in $\mathbf{S}$ are causally unconnected to any other variables in $\mathbf{V}$. In that case, when we depict a DAG in a figure we will omit the variables in $\mathbf{S}$, and edges which have an endpoint in $\mathbf{S}$.

---

[1]For readers interested in following the examples in some detail, under our assumptions, the conditional independence relations entailed by a DAG are given by the d-separation relation. See Pearl 1988 for details.

For a given DAG G, and a division of the variable set $\mathbf{V}$ of G into observed ($\mathbf{O}$), selection ($\mathbf{S}$), and latent ($\mathbf{L}$) variables, we will write $G(\mathbf{O,S,L})$. We assume that the only conditional independence relations that can be tested are $\mathbf{X}\perp\!\!\!\perp\mathbf{Z}|\mathbf{Y}\cup(\mathbf{S}=\mathbf{1})$, where $\mathbf{X}$, $\mathbf{Z}$, and $\mathbf{Y}$ are subsets of $\mathbf{O}^2$; we will call this the set of *observable* conditional independence relations. If $\mathbf{X}$, $\mathbf{Y}$, and $\mathbf{Z}$ are included in $\mathbf{O}$, and $\mathbf{X}\perp\!\!\!\perp\mathbf{Z}|\mathbf{Y}\cup(\mathbf{S}=\mathbf{1})$, then we say it is an observed conditional independence relation. Cond($\mathbf{O}$) is a set of conditional independence relations among variables in $\mathbf{O}$. A DAG $G(\mathbf{O,S,L})$ entails Cond($\mathbf{O}$) just when for each $\mathbf{X}$, $\mathbf{Y}$, and $\mathbf{Z}$ included in $\mathbf{O}$, it entails $\mathbf{X}\perp\!\!\!\perp\mathbf{Z}|\mathbf{Y}\cup(\mathbf{S}=\mathbf{1})$ if and only if $\mathbf{X}\perp\!\!\!\perp\mathbf{Z}|\mathbf{Y}$ is in Cond($\mathbf{O}$). (Note that $\mathbf{O}$ is the same for all DAGs that entail Cond($\mathbf{O}$), but $\mathbf{S}$ is not. This is because we are allowing some DAGs to collapse multiple selection variables into a single selection variable.) We will simply write Cond when the context makes it clear what $\mathbf{O}$ is. However, there may be many different DAGs that entail exactly the same Cond; the set of all DAGs that entail a given Cond we will call Equiv(Cond).

Imagine now that a researcher does not know what the correct causal DAG is, but can determine what Cond is, perhaps by performing hypothesis tests of conditional independence relations on the selected subpopulation. (As we will see later, because many of the members of Cond entail other members of Cond, only a fraction of the membership of Cond actually need be tested.) From this information alone, and the Causal Markov Assumption, the Causal Faithfulness Assumption, and the Population Inference Assumption, the most he or she could conclude is that the true causal DAG is either $G_1(\mathbf{O,S_1,L_1})$ or $G_2(\mathbf{O,S_2,L_2})$ or $G_3(\mathbf{O,S_3,L_3})$ or $G_4(\mathbf{O,S_4,L_4})$, etc., i.e. the true causal DAG is some member of Equiv(Cond). If Equiv(Cond) is large, then this information by itself is not very interesting, unless the members of Equiv(Cond) all share some important features in common. However, as the following examples show, in some cases the members of Equiv(Cond) *do* share important features in common.

[2]We are looking only at the subpopulation where all of the variables in $\mathbf{O}$ have recorded values. A more complicated story could be told if when considering conditional independence relations among $\mathbf{X}$, $\mathbf{Z}$, and $\mathbf{Y}$ we looked only at subpopulations where just the selection variables corresponding to $\mathbf{X}$, $\mathbf{Z}$, and $\mathbf{Y}$ have values equal to 1.

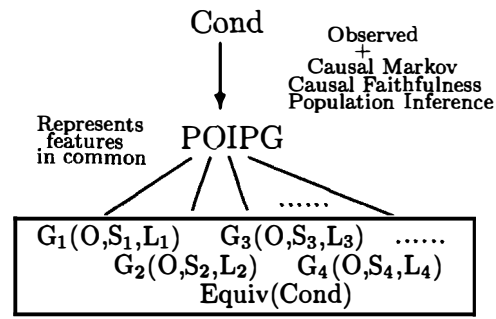

Figure 3: General strategy

Our strategy, which is a generalization of the strategy without selection bias described in Spirtes et al. (1993), will be to construct from Cond a graphical object called a *partially oriented inducing path graph* (POIPG), using the Causal Markov Assumption, the Causal Faithfulness Assumption, and the Population Inference Assumption. The POIPG (described in more detail in the examples and the Appendix) represents certain features that all of the DAGs in Equiv(Cond) share in common.[3] From the constructed POIPG it is sometimes possible to infer that all of the DAGs in Equiv(Cond) share some other interesting features in common, e.g. they might all contain a directed path from A to B. If this is the case, then although from Cond we cannot tell exactly which DAG in Equiv(Cond) is the true causal DAG, because we know that all of the DAGs in Equiv(Cond) contain a directed path from A to B we can reliably conclude that in the true causal DAG A is a (possibly indirect) cause of B. This strategy is represented schematically in Figure 3. In the following examples we will apply this strategy to particular sets of observed conditional independence relations, and show what features of DAGs can be reliably inferred.

## 4.1   Example 1

We will start out with a very simple example, in which the set of observed conditional independence relations is not very informative. (For simplicity, in all of the following examples we assume that all of the variables in $\mathbf{S}$ take on the same value, and hence can be represented by a single variable S.) Let $\mathbf{O} = \{\mathbf{A,B}\}$. and the set Cond1 of observed conditional independence relations is empty, i.e. Cond1 = $\emptyset$. We now want to find out what DAGs are in Equiv(Cond1). Let $\mathbf{V}$ be a set of causally sufficient variables. The simplest example of such a DAG is when $\mathbf{V} = \mathbf{O} = \{\mathbf{A,B}\}$ and there is *no selection bias*. (That $\mathbf{V} = \mathbf{O}$ and there is no selection bias is typically either an assumption or comes from background knowledge, since it is not in general possible to definitively confirm these condi-tions from the data alone.) Under these assumptions there are exactly two DAGs that entail Cond1, labeled (i) and (ii) in Figure 4. In general, when there are no latent variables and no selection bias, there is an edge between A and B if and only if for all subsets $\mathbf{X}$ of $\mathbf{O}\backslash\{A,B\}$, A and B are dependent given $\mathbf{X} \cup (\mathbf{S}=\mathbf{1})$.

[3]POIPGs are generalizations of structures described (but not named) in Verma and Pearl 1991, and share some features in common with the representation scheme used in Wermuth, Cox, and Pearl 1994.

Now suppose that there are latent variables but no selection bias. Then, if we do not limit the number of latent variables in a DAG, there are an infinite number of DAGs that entail Cond1, many of which do not contain an edge between A and B. Two such DAGs are shown in (iii) and (vi) of Figure 4. The examples in (iii) and (vi) of Figure 4 show that when there are latent variables it is not the case that there is an edge between A and B if and only if for all subsets $\mathbf{X}$ of $\mathbf{O}\backslash\{A,B\}$, A and B are dependent given $\mathbf{X}\cup\mathbf{S}$. (Recall that if there is no selection bias that A and B are dependent given $\mathbf{X} \cup \mathbf{S}$ if and only if A and B are dependent given $\mathbf{X}$.)

Finally, let us consider the case where there is selection bias. Examples of such DAGs in Equiv(Cond1) are shown in (iv) and (v) of Figure 4.

The DAGs in Equiv(Cond1) seem to have little in common, particularly when there is the possibility of both latent variables and selection bias. While there is a great variety of DAGs in Equiv(Cond1), it is not the case that every DAG is in Equiv(Cond1). For example, a DAG $G(\mathbf{O},\mathbf{S},\mathbf{L})$ with no edges at all is not in Equiv(Cond1).

In Equiv(Cond1), for each subset $\mathbf{X}$ of $\mathbf{O}$, A and B are dependent given $\mathbf{X} \cup \mathbf{S}$ (in this example $\mathbf{X}$ is just the empty set). In general, for any $\mathbf{O}$, there is a simple graphical characterization of this condition. A DAG $G(\mathbf{O},\mathbf{S},\mathbf{L})$ entails that for each subset $\mathbf{X}$ of $\mathbf{O}$, A and B are dependent given $\mathbf{X} \cup \mathbf{S}$ if and only if a certain kind of undirected path between A and B exists in $G(\mathbf{O},\mathbf{S},\mathbf{L})$. This undirected path is called an inducing path. U is an inducing path between A and B in $\mathbf{O}$ in DAG $G(\mathbf{O},\mathbf{L},\mathbf{S})$ if and only if U is an acyclic undirected path between A and B such that every collider on U has a descendant in $\{A,B\} \cup \mathbf{S}$, and no non-collider on U except for the endpoints is in $\mathbf{O} \cup \mathbf{S}$. (A vertex V is a collider on an undirected path U if and only if two adjacent edges on U are into V, i.e. there exist variables X and Y on U, such that $X \rightarrow V \leftarrow Y$ is a subpath of U.)

The orientation of an inducing path with endpoints X and Y is determined by whether or not the edges on the inducing path containing X and Y have arrowheads at X and Y respectively. There are four possible orientations of inducing paths between A and B, all of which are shown in Figure 4. There are inducing paths that are out of A and out of B (e.g. the inducing path $A \rightarrow S \leftarrow B$ in (iv)), inducing paths that are out of A and into B (e.g. the inducing path $A \rightarrow B$ in (i) and the inducing path $A \rightarrow S \leftarrow T \rightarrow B$ in (v)), inducing paths that are out of B and into A (e.g. the inducing path $B \rightarrow A$ in (ii)), and inducing paths that are into

A and into B (e.g. the inducing path $A \leftarrow T \rightarrow B$ in (iii) and the inducing paths $A \leftarrow T \rightarrow B$ and $A \leftarrow U \rightarrow B$ in (vi)). Hence, if Cond1 is observed, it is possible to tell that there is an inducing path between A and B, but not what the orientation of the inducing path is. We represent this information in a partially oriented inducing path graph with the edge A o—o B. The fact that A and B are adjacent in the POIPG means that there is an inducing path between A and B; the "o" on each end of the edge means that we cannot tell what the orientation of the inducing path is. The existence and orientation of inducing paths is typically not particularly interesting information about the DAGs in Equiv(Cond1). Can we tell anything more interesting about the causal relationship between A and B from the POIPG? In this case, the answer is no; however, the next example shows a case where more interesting conclusions can be drawn.

### 4.2 Example 2

Let $\mathbf{O} = \{A,B,C,D\}$ and Cond2 $= \{D\perp\!\!\!\perp\{A,B\}|\{C\}, A\perp\!\!\!\perp B\}$ and all of the other conditional independence relations entailed by these. The only DAG in Equiv(Cond2) with no latent variables and no selection bias is (i) in Figure 5.

Now suppose that we consider DAGs with latent variables so $\mathbf{V} \neq \mathbf{O}$, but there is no selection bias. In that case if there is no upper limit to the number of latent variables allowed, then there are an infinite number of DAGs in Equiv(Cond2), several of which are shown in (ii), (iii) and (iv) of Figure 5.

Suppose that we now consider DAGs with selection bias. (v) and (vi) of Figure 4.2 are examples of DAGs that are in Equiv(Cond2) and have selection bias.

Is there anything that all of the DAGs in Figure 4.2 have in common? There are no inducing paths between the pairs $\langle A,D\rangle$, $\langle B,D\rangle$ or $\langle A,B\rangle$ in any of the DAGs in Equiv(Cond2) because for each of these pairs there is a subset $\mathbf{X}$ of $\mathbf{O}$ such that they are independent conditional on $\mathbf{X} \cup \mathbf{S}$. This is represented in the POIPG by the lack of edges between A and D, between B and D, and between A and B. Because for each subset $\mathbf{X}$ of $\mathbf{O}$, A and C are dependent conditional on $\mathbf{X} \cup \mathbf{S}$, we can conclude that there is an inducing path between A and C. Moreover, in the DAGs in Figure 5 while some of the inducing paths are out of A, and others are into A, note that they are all into C. It can be shown that *all* of the inducing paths between A and C in *all* of the DAGs in Equiv(Cond2) are into C. In the POIPG representing Equiv(Cond2) we represent this by A o→ C. A and C are adjacent in the POIPG because there is an inducing path between A and C. The "o" on the A end of the edge means we cannot tell the orientation of the A end of the inducing path between A and C; the ">" on the C end of the edge means that all of the inducing paths between A and C in all of the DAGs in Equiv(Cond2) are into C. It is also the case that all of the DAGs in Figure 5 have inducing paths between C and D that are out of C and into D. It can be shown that all of the inducing paths between C and D in all of the DAGs in Equiv(Cond2) are out of C and into D. These facts are represented in the POIPG by the edge between C and D having a ">" at the D end and a "–" at the C end.

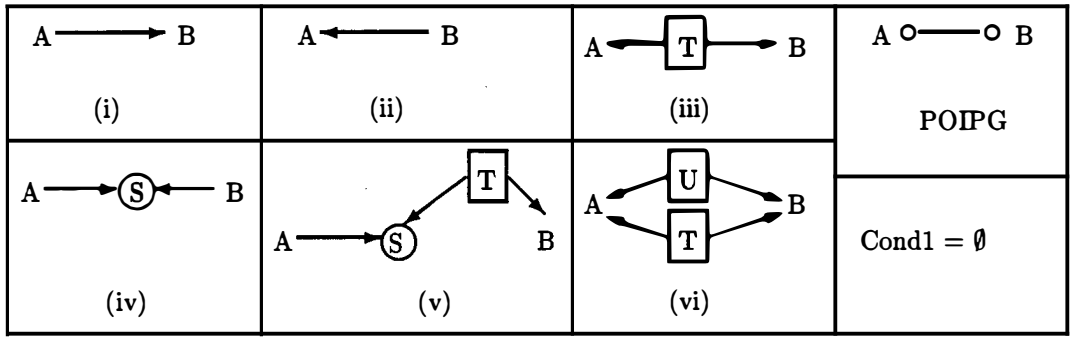

Figure 4: Some members of Equiv(Cond1)

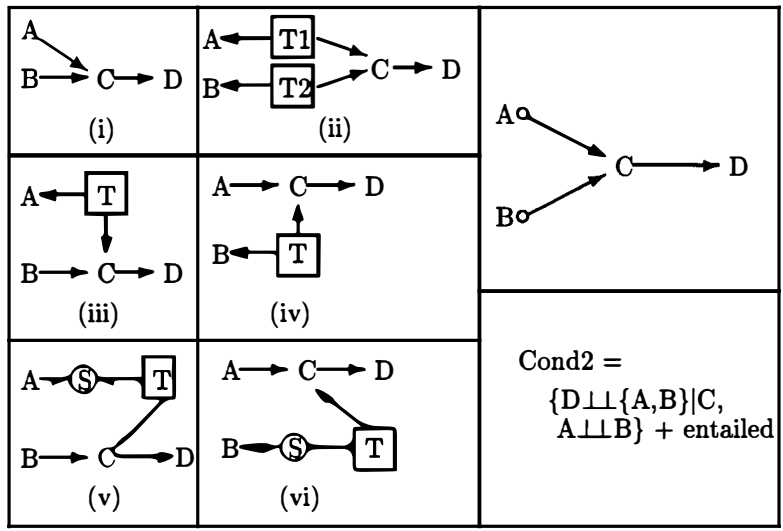

Figure 5: Some members of Equiv(Cond2)

It can be shown from the fact that the POIPG contains the edge C → D that there is a directed path from C to D that does not contain any member of **S** in every DAG in Equiv(Cond2), i.e. C is a (possibly indirect) cause of D. It can also be shown that the POIPG of Equiv(Cond2) entails that there are no directed paths from D to A or from D to B in any of the DAGs in Equiv(Cond2), i.e. D is not a cause (either direct or indirect) of either A or B.

### 4.3     Example 3

Finally, consider an example in which **O** = {A,B,C,D}, and Cond3 = {D⊥⊥{A,B}, A⊥⊥{C,D}} and all of the other conditional independence relations entailed by these. There is no DAG in Equiv(Cond3) in which **V** = **O**. Hence we can conclude that each DAG in Equiv(Cond3) either contains a latent variable. (i) and (ii) of Figure 6 are examples of DAGs with latent variables in Equiv(Cond3). Note that in each of them, there is a latent common cause of B and C, C is not a descendant of B, and B is not a descendant of C. As long as there is no selection bias, these properties can be shown to hold of any DAG in Equiv(Cond3).

Suppose now that we also consider DAGs with selection bias. (iii) of Figure 6 is an example of a DAG with selection bias that is in Equiv(Cond3). Note that the inducing paths between B and C are into B and into C in every DAG in Figure 6; this can be shown to be the case for every inducing path between B and C in every DAG in Equiv(Cond3). Hence in the POIPG we have an edge B ↔ C. Note that (iii) in Figure 6 does not contain a latent common cause of C and B. However, in each of the DAGs in Equiv(Cond3) C is not a descendant of B, and B is not a descendant of C; these properties can be shown to hold of any DAG in Equiv(Cond3), even when there are latent variables and selection bias. Hence if the conditional independence relations in Cond3 are ever observed, it can be reliably concluded that even though there may be latent variables and selection bias, and regardless of the causal connections of the latent variables and selection variables to other variables, in the causal DAG that generated Cond3, B is not a direct or indirect cause of C and C is not a direct or indirect cause of B.

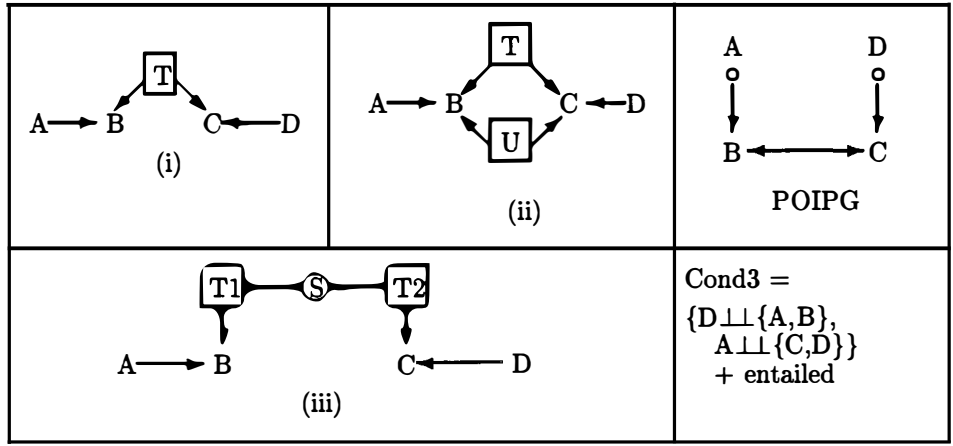

Figure 6: Some members of Equiv(Cond3)

## 5 AN ALGORITHM FOR CONSTRUCTING POIPGs

We have seen that a POIPG contains valuable information about the causal relationships between variables. However, the number of observable conditional independence relations grows exponentially with the number of members of **O**. In addition, some of the independence relations are conditional on large sets of variables, and often these cannot be reliably tested on reasonable sample sizes. Is it possible to construct a POIPG?

The FCI algorithm (Spirtes et al. 1993) constructs correct POIPGs, under the Causal Markov Assumption, the Causal Faithfulness Assumption, the assumptions of no selection bias, and that independence relations can be reliably tested. If the possibility of selection bias is allowed, the algorithm described in Spirtes et al. (1993) still gives the correct output, but the conclusions that one can draw from the POIPG are the slightly weaker ones described in the examples and the Appendix to this paper. In the worst case the FCI algorithm is exponential time in the number of variables, even when the maximum number of vertices any given vertex is adjacent to is held fixed. However, on simulated data the algorithm can often be run on up to 100 variables provided the true graph is sparse. This is because it is not necessary to examine the entire set of observable conditional independence relations; many conditional independence relations are entailed by other conditional independence relations. The FCI algorithm relies on this fact to test a relatively small set of conditional independence relations, and test independence relations conditional on as few variables as possible.

## 6 APPENDIX

A *directed graph* is an ordered pair $\langle V,E \rangle$ of vertices **V** and edges **E**, where each edge is an ordered pair of distinct vertices. The ordered pair $\langle A,B \rangle$ is an edge *from* A to B and is *out of* A and *into* B. If there is an edge from A to B then A is a *parent* of B. If there is an edge from A to B or from B to A, then A and B are *adjacent*. An *undirected path* between $X_1$ and $X_n$ in directed graph G is a sequence of vertices $\langle X_1, \ldots, X_n \rangle$ such that for $1 \leq i < n$, $X_i$ and $X_{i+1}$ are adjacent in G. $X_i$ is a *collider* on undirected path U if and only if there are edges from $X_{i-1}$ and $X_{i+1}$ to $X_i$ on U. An undirected path U between $X_1$ and $X_n$ is *into* (*out of*) $X_1$ if and only if the edge on U between $X_1$ and $X_2$ is into (out of) $X_1$, and similarly for $X_n$. A *directed path* from $X_1$ to $X_n$ in directed graph G is a sequence of vertices $\langle X_1, \ldots, X_n \rangle$ such that for $1 \leq i < n$, there is a directed edge from $X_i$ to $X_{i+1}$ in G. (A path may consist of a single vertex.) A path (directed or undirected) is *acyclic* if and only if it contains no vertex more than once. A directed graph is *acyclic* if every directed path in directed graph G is acyclic. A is an *ancestor* of B and B is a *descendant* of A if and only if A = B or there is a directed path from A to B.

Theorem 1 generalizes results of Verma and Pearl (1991).

**Theorem 1** *A DAG G(**O,S,L**) entails that for all subsets* **X** *of* **O**, *A is dependent on B given* (**X** ∪ **S**)\\$\{A, B\}$ *if and only if there is an inducing path between A and B.*

A DAG G′(**O,S′,L′**) is in Equiv(G(**O,S,L**)) if and only if for all disjoint subsets **X**, **Y**, and **Z** of **O**, G(**O,S,L**) entails $\mathbf{X} \perp\!\!\!\perp \mathbf{Z} | \mathbf{Y} \cup (\mathbf{S} = 1)$ if and only if G′(**O,S′,L′**) entails $\mathbf{X} \perp\!\!\!\perp \mathbf{Z} | \mathbf{Y} \cup (\mathbf{S}' = 1)$.

In the following definition we use the symbol "*" as a wild-card symbol to denote any kind of edge endpoints in a partially oriented inducing path graph π; "*" itself never occurs in π. π is a *partially oriented inducing* path graph of directed acyclic graph G(**O**,**S**,**L**) if and only if

(i) the set of vertices in $\pi$ equals **O**;

(ii) if there is any edge between A and B in $\pi$, it is one of the following kinds: A $\rightarrow$ B, A o$\rightarrow$ B, A o$-$o B, or A $\leftrightarrow$ B

(iii) there is an edge between A and B in $\pi$ if and only if for each $G_1(\textbf{O},\textbf{S}',\textbf{L}')$ in Equiv(G(**O**,**S**,**L**)) there is an inducing path between A and B;

(iv) there is at most one edge between any pair of vertices A and B in $\pi$;

(v) if A $-$* B is in $\pi$, then for each $G_1(\textbf{O},\textbf{S}',\textbf{L}')$ in Equiv(G(**O**,**S**,**L**)) every inducing path between A and B is out of A;

(vi) if A *$\rightarrow$ B is in $\pi$, then for each $G_1(\textbf{O},\textbf{S}',\textbf{L}')$ in Equiv(G(**O**,**S**,**L**)) every inducing path between A and B is into B;

(vii) if A *$-$* B *$-$* C is in $\pi$, then for each $G_1(\textbf{O},\textbf{S}',\textbf{L}')$ in Equiv(G(**O**,**S**,**L**)) there is no pair of inducing paths U and V such that U is between A and B and into B, and V is between B and C and into B.

(We sometime write A $\leftarrow$ B for B $\rightarrow$ A, and A $\leftarrow$o B for B o$\rightarrow$ A.)

Informally, a directed path in a POIPG is a path that contains only "$\rightarrow$" edges pointing in the same direction.

**Theorem 2** *If $\pi$ is a partially oriented inducing path graph, and there is a directed path U from A to B in $\pi$, then in every DAG G(**O**,**S**,**L**) with POIPG $\pi$ there is a directed path from A to B, and A has no descendant in **S**.*

**Theorem 3** *If $\pi$ is a partially oriented inducing path graph and A $\leftrightarrow$ B in $\pi$, then there is a latent variable and no directed path from A to B and no directed path from B to A in any DAG G(**O**,**S**,**L**) with POIPG $\pi$.*

A *semi-directed path* from A to B in a partially oriented inducing path graph $\pi$ is an undirected path acyclic U between A and B in which no edge contains an arrowhead pointing towards A, (i.e. there is no arrowhead at A on U, and if X and Y are adjacent on the path, and X is between A and Y on the path, then there is no arrowhead at the X end of the edge between X and Y). Theorems 4, 5, and 6 give information about what variables appear on causal paths between a pair of variables A and B, i.e. information about how those paths could be blocked.

**Theorem 4** *If $\pi$ is a partially oriented inducing path graph, and there is no semi-directed path from A to B in $\pi$ that contains a member of **C**, then every directed path from A to B in every DAG G(**O**,**S**,**L**) with POIPG $\pi$ that contains a member of **C** also contains a member of **S**.*

**Theorem 5** *If $\pi$ is a partially oriented inducing path graph, and there is no semi-directed path from A to B in $\pi$, then every directed path from A to B in every DAG G(**O**,**S**,**L**) with POIPG $\pi$ contains a member of **S**.*

**Theorem 6** *If $\pi$ is a partially oriented inducing path graph, and every semi-directed path from A to B contains some member of **C** in $\pi$, then every directed path from A to B in every DAG G(**O**,**S**,**L**) with POIPG $\pi$ contains a member of **S** $\cup$ **C**.*

## Acknowledgments

We would like to thank two anonymous referees for helpful comments on an earlier draft of this paper, and Clark Glymour and Greg Cooper for helpful conversations. Research for this paper was supported by the Office of Navel Research grant ONR #N00014-93-1-0568.

## References

Bollen, K. (1989). *Structural Equations with latent variables.* New York: Wiley.

Cooper, G. (1995). Causal discovery from data in the presence of selection bias. In *Preliminary papers of the fifth international workshop on Artificial Intelligence and Statistics*, Fort Lauderdale, FL, pp. 140–150.

Lauritzen, S., A. Dawid, B. Larsen, and H. Leimer (1990). Independence properties of directed Markov fields. *Networks 20.*

Meek, C. (1995). Strong completeness and faithfulness in Bayesian networks. In *Proceedings of Uncertainty in Artificial Intelligence.* To appear.

Pearl, J. (1988). *Probabilistic Reasoning in Intelligent systems.* San Mateo: Morgan-Kaufmann.

Spirtes, P., C. Glymour, and R. Scheines (1993). *Causation, Prediction, and Search.* Springer-Verlag.

Verma, T. and J. Pearl (1991). Equivalence and synthesis of causal models. In *Proceedings of the Sixth Conference in Art. Int.*, Mountain View, CA, pp. 220–227. Association for Uncertainty in AI.

Wermuth, N., D. Cox, and J. Pearl (1994). Explanations for multivariate structures derived from univariate recursive regressions. Technical Report Report 94-1, University of Mainz.

Wright, S. (1934). The method of path coefficients. *Annals of Mathematical Statistics 5.*